# Steering Responsible AI: A Case for Algorithmic Pluralism

Dr. Stefaan G. Verhulst

**Working Paper – Open for Review**

## Abstract

As artificial intelligence (AI) continues to grow in scope and sophistication, it is likely to play an increasingly central role in shaping how citizens interact with reality. In one version of the not-so-distant future, AI apps and bots will increasingly displace the dominant mediating platforms of today, becoming the new mediators of our online lives. Whether or not such a scenario does eventually unfold, its very possibility—likelihood?—strongly suggests that we should be asking the same questions of AI that we have grown accustomed to asking of our existing mediators. In particular, we need to puncture the "fallacy of AI neutrality - represented by the mistaken belief that AI systems can be designed in an inherently unbiased and neutral manner. In this paper, I examine questions surrounding AI neutrality through the prism of existing literature and scholarship about mediation and media pluralism. Such traditions, I argue, provide a valuable theoretical framework for how we should approach the (likely) impending era of AI mediation. In particular, I suggest examining further the notion of algorithmic pluralism. Contrasting this notion to the dominant idea of algorithmic transparency, I seek to describe what algorithmic pluralism may be, and present both its opportunities and challenges. Implemented thoughtfully and responsibly, I argue, Algorithmic or AI pluralism has the potential to sustain the diversity, multiplicity, and inclusiveness that are so vital to democracy.

# Introduction

As artificial intelligence (AI) continues to grow in scope and sophistication, it is likely to play an increasingly central role in shaping how citizens interact with reality. In one version of the not-so-distant future, AI apps and bots will increasingly displace the dominant mediating platforms of today, becoming the new mediators of our online lives. We are already seeing hints of this possibility in suggestions that (generative) AI could eventually dislodge Google from its perch as the leading search engine; in much the same way, other AI use cases and platforms could display the other portals that today shape the contours of our increasingly online experience.

Whether or not such a scenario does eventually unfold, its very possibility—*likelihood*?—strongly suggests that we should be asking the same questions of AI that we have grown accustomed to asking of our existing mediators. In particular, we need to puncture what Nora Khan, in an essay titled "Seeing, Naming, Knowing[1]," labels the "fallacy of neutrality." The fallacy of neutrality is represented by the mistaken belief that AI systems can be designed in an inherently unbiased and neutral manner. However, as Khan points out, these systems are designed and trained by humans, who inevitably bring their own biases and perspectives to the process. In addition, as is now well documented, they are built using data that is itself riddled with biases, further undermining the purported neutrality of AI models.

In what follows, I examine questions surrounding AI neutrality through the prism of existing literature and scholarship about mediation and media pluralism[2] [3]. Such traditions, I argue, provide a valuable theoretical framework for how we should approach the (likely) impending era of AI mediation. In particular, I suggest to examine further the notion of algorithmic pluralism. Contrasting this notion to the idea of algorithmic transparency, I seek to describe what algorithmic pluralism may be, and present both its opportunities and challenges. Implemented thoughtfully and responsibly, I argue, Algorithmic or AI pluralism has the potential to sustain the diversity, multiplicity, and inclusiveness that are so vital to democracy.

---

[1] Kahn, Nora. "Seeing, Naming, Knowing. The Brooklyn Rail. (March, 2019)
[2] Labaran, Musa Adamu. "One World, One Network: Revisiting Digital Dichotomy Theory of the Media." *Konfrontasi: Jurnal Kultural, Ekonomi dan Perubahan Sosial* 10, no. 3 (2023): 159-166.
[3] Sartori, Laura, and Giulia Bocca. "Minding the gap (s): public perceptions of AI and socio-technical imaginaries." AI & society 38, no. 2 (2023): 443-458.

## Mediation, Pluralism and AI

The idea of mediation is, of course, not new. The notion has for decades served as an organizing principle in sociological, cultural and legal approaches to the governance of media and mass communication.[4] Historically, scholars have drawn attention to the media's role as a gatekeeper[5] that plays a critical role in shaping our perceptions—and experiences--of society.[6] Unpacking McLuhan's famous dictum that "the medium is the message," Chakravorty, for instance, argues that the construction (or imposition) of meaning by the medium points to the mediating function performed by various channels of mass communication.[7]

Mediation is about framing.[8] In his widely quoted definition, Todd Gitlin defines frames as "principles of selection, emphasis, and presentation composed of little tacit theories about what exists, what happens, and what matters".[9] Media frames, in other words, determine what parts of reality we notice. They shape our perceptions of reality, and they provide a way to "understand" events, building on existing "frames of reference" and embedded knowledge.[10]

Media scholars—and policymakers—have long recognized that, when media entities consolidate or concentrate, there's a risk of power asymmetry where a singular narrative or dominant framing overshadows others. This is not just seen as a threat to the richness of intellectual discourse, but also to the foundation of democratic societies, where decision-making and public participation benefit from a plethora of insights and perspectives. A marketplace of ideas, where various interpretations and viewpoints are freely exchanged, is believed to be fundamental to a thriving democracy, as it ensures that the public is not swayed by a single dominant narrative. In this light, pluralism isn't just an ideal but a necessity. Its importance is only heightened by Khan's "fallacy of neutrality"—i.e., the recognition that no single source of information or mediator can present a "true" or unbiased point of view.

---

[4] *Media and Mediation*, ed. Bernard Bel, Jan Brouwer, Biswajit Das, Vibodh Parthasarthi, Guy Poitevin (New Delhi: SAGE India, 2005).

[5] Lewin, Kurt. "Frontiers in Group Dynamics: II. Channels of Group Life; Social Planning and Action Research," *Human Relations* 1, no. 2 (1947), https://doi.org/10.1177/001872674700100201 ; White, David Manning "The 'Gate Keeper': A Case Study in the Selection of News," *Journalism and Mass Communication Quarterly* 27, no. 4 (1950), https://doi.org/10.1177/107769905002700403.

[6] Baudrillard, Jean. *Simulations*, trans. Paul Foss, Paul Patton, Philip Beitchman (New York: Semiotext(e), 1983).

[7] Chakravorty,Swagato. "mediation," *The Chicago School of Media Theory* (blog), 2010, https://lucian.uchicago.edu/blogs/mediatheory/keywords/mediation/.

[8] Erving, Goffman. "Frame analysis: An essay on the organization of experience." (1974).

[9] Gitlin, Todd. *The whole world is watching: Mass media in the making and unmaking of the new left*. Univ of California Press, 2003.

[10] Cappella, Joseph N., and Kathleen Hall Jamieson. Spiral of cynicism: The press and the public good. Oxford University Press, 1997.

Recognizing that no mediation is truly neutral, policymakers have developed approaches to actively champion diversity in media sources and representation. In certain contexts, public sponsorship of (public) media outlets has been deemed to be crucial to preserve this diversity, ensuring that less commercially viable, yet socially important, perspectives find a platform. In other contexts, as I have documented and assessed elsewhere[11], media pluralism has been fostered through regulatory measures to prevent the concentration of media ownership, such as antitrust laws and ownership rules. Content quotas have also played a role, mandating a certain amount of domestic or culturally specific programming to support local narratives. Finally, media literacy programs have been instrumental in encouraging critical engagement with media, fostering a populace capable of consuming a diverse range of media critically.

These issues of mediation and power asymmetries, always important, have recently risen to prominence with the "datafication"[12] of society and the growing importance of algorithms and AI. AI is already playing a major role in a variety of everyday functions and decisions, including consumption and content recommendations, and much more. The algorithms that power these functions are emerging as de facto mediators for the data age. Their role is all the more important — and potentially pernicious —given the intersection of the data age and power.

Dominant or monopolistic mediators (such as, large tech companies), with privileged or asymmetric access to data and the ability to use that data,[13] are playing an increasingly central role in shaping our political and cultural discourse, and in determining our norms and values. As an example, Facebook has played an outsized role in mediating civic participation by algorithmic curation and platform affordances.[14] Similarly, Twitter or now X's algorithmic content curation[15] tends to create a partial bias towards partisan echo chambers, rather than promoting a broader exposure to mainstream sources. In other forms of media consumption, such as music listening habits, researchers[16] find that users increasingly find algorithmic personalization to be fundamentally impersonal, and there is a spectrum between algorithmically led listening habits and user-led interactions with new music in platforms such as Spotify. These examples demonstrate that there is an increased reliance on monopolistic mediators in shaping political and cultural discourse, although there are some limits to the extent to which algorithms can influence individual preferences.

---

[11] Verhulst, Stefaan G. European Responses to Media Ownership and Pluralism - Introduction 16 Cardozo Arts & Ent. L.J. 421 (1998)
[12] Verhulst, Stefaan G. "Operationalizing digital self-determination." Data & Policy 5 (2023): e14.
[13] Verhulst, Stefaan G. "The ethical imperative to identify and address data and intelligence asymmetries." AI & Society (2022): 1-4.

The importance of exposing the fallacy of neutrality is therefore arguably more important now than ever before. As AI increasingly shapes our world and our perceptions of what's important and real, it is vital that we acknowledge the bias of such representations and hold on to the diversity and multi-stranded nature of reality.

## Toward Algorithmic Pluralism

Today most discussions about algorithmic bias involve pushing for greater transparency. While the push for transparency is worthy and important, it must exist alongside efforts to introduce choice in the way citizens interact with AI algorithms and models. As a movement, algorithmic pluralism begins by openly acknowledging the biases inherent in the design and training of AI systems, allowing different audiences to effectively choose the biases they engage with. The idea is that there should not be a one-size-fits-all approach to AI; instead, different algorithms could be designed to reflect different perspectives, preferences and values,[17] thereby providing a more nuanced and diverse approach to AI.

---

[14] Papa, Venetia, and Nikandros Ioannidis. "Reviewing the impact of Facebook on civic participation: The mediating role of algorithmic curation and platform affordances." The Communication Review 26, no. 3 (2023): 277-299.

[15] Bandy, Jack, and Nicholas Diakopoulos. "More accounts, fewer links: How algorithmic curation impacts media exposure in Twitter timelines." Proceedings of the ACM on Human-Computer Interaction 5, no. CSCW1 (2021): 1-28.

[16] Freeman, Sophie, Martin Gibbs, and Bjorn Nansen. "Personalised But Impersonal: Listeners' Experiences of Algorithmic Curation on Music Streaming Services." In Proceedings of the 2023 CHI Conference on Human Factors in Computing Systems, pp. 1-14. 2023.

[17] Verhulst, S., and Mona Sloane. "Realizing the potential of AI localism." Project Syndicate 7 (2020).

Algorithmic pluralism remains a nascent concept–promising but still taking shape, and not without its share of challenges. While ultimate use cases are far wider, they originate with a recognition of the limitations of social media platforms, particularly what Eli Pariser has called the "filter bubbles"[18] that reinforce biases and spread misinformation. Cass Sunstein has also referred to this as the "Daily Me" effect.[19] Such bubbles, it is increasingly clear, pose a number of risks, including to mental health (particularly of minors), democracy, and public discourse.

The push for algorithmic pluralism is evident in steps taken both by regulators and private companies. The EU's Digital Services Act (DSA), for instance, requires social media platforms to explain how information in feeds is presented and to conduct audits on the manner in which their algorithms affect democracy and mental health. In addition, the framework also requires platforms to present users with the choice of at least one algorithm that doesn't present information based on "behavioral profiles[20]." Bluesky[21], a public benefit company created in 2021, represents similar efforts by a private sector enterprise. Spun out of Twitter, the company begins from the premise that "algorithms to help people sort through information must evolve rapidly" and states a goal of creating "a system that enables composable, customizable feed generation[22]" through a marketplace of algorithms open to third party developers. Writing in the New York Times, Julia Angwin explains how this ability to select feeds works in practice:

> *On my Bluesky feed, I often toggle among feeds called Tech News, Cute Animal Pics, PositiviFeed and my favorite, Home+, which includes "interesting content from your extended social circles." Some of them were built by Bluesky developers, and others were created by outside developers. All I have to do is go to My Feeds and select a feed from a wide menu of choices, including MLB+, a feed about baseball; #Disability, one that picks up keywords related to disability; and UA fundraising, a feed of Ukrainian fund-raising posts.[23]*

---

[18] Pariser, Eli. *The filter bubble: How the new personalized web is changing what we read and how we think*. Penguin, 2011.
[19] Sunstein, Cass. *# Republic: Divided democracy in the age of social media*. Princeton University Press, 2018.
[20] Hello World. "Understanding the Digital Services Act" The Markup. (April 30, 2022)
[21] https://blueskyweb.xyz/
[22] Graber, Jay. Algorithmic Choice. Bluesky. (March 30, 2023) https://blueskyweb.xyz/blog/3-30-2023-algorithmic-choice
[23] Angwin, Julia. What if You Knew What You Were Missing on Social Media? The New York Times. (August 17, 2023) https://www.nytimes.com/2023/08/17/opinion/social-media-algorithm-choice.html

# Opportunities—and challenges—of Algorithmic Pluralism

Algorithmic pluralism represents a potentially significant transformation in the way citizens interact with social media platforms and consume information. The effects on society, culture and politics could be profound. As noted, however, the push for greater pluralism is nascent, and the concept remains relatively undeveloped, both as an idea and in its technical feasibility. In this section, we explore some opportunities offered by algorithmic pluralism, as well as some remaining questions and challenges:

## Opportunities:

- Promotes choice: Efforts to regulate media platforms, while often laudable in their intentions, do raise the possibility of censorship and government control—which, in turn, can stifle innovation and limit free choice. As Damien Tambini has argued, systems that allow for some kind of algorithmic pluralism have the virtue of maintaining the "core values and liberties that define democracy[24]." By in effect regulating choice, algorithmic pluralism offers a third or middle way between the undeniable need for greater regulation and the perils of over-regulation.

- Enhances democracy: One of the chief concerns over current algorithmic systems concerns their erosion of democracy and civic discourse. These concerns, intimately linked to the issue of filter bubbles[25], have risen to prominence in recent years, particularly with the global rise of illiberal democracy and its accompanying problems of misinformation and hate speech. By increasing the diversity of views in the marketplace, and by enabling citizens to step out of their personalized echo chambers, algorithmic choice holds the potential to lead to a healthier discourse and greater levels of trust in society.

- Promotes a healthy marketplace and encourages innovation: One of the ironies of the modern data and media ecology is that unrestricted markets have in effect led to an erosion of freedom in markets. By ensuring diversity, algorithmic pluralism in effect leads to a more robust media marketplace, eroding the power of monopolistic platforms and reducing the asymmetric influence of today's market leaders. In so doing, the value of pluralism also encourages greater innovation—specifically with regard to algorithms themselves, but more generally in the broader technical ecology.

---

[24] Tambini, Damian. "Media regulation and system resilience in the age of information warfare." The World Information War: Western Resilience, Campaigning, and Cognitive Effects (2021).
[25] Spohr, Dominic. "Fake news and ideological polarization: Filter bubbles and selective exposure on social media." Business information review 34, no. 3 (2017): 150-160.

- Transparency and Trust: Trust is key to innovation and a robust democracy. In recent years, a rising sense of deceptive practices and algorithmic manipulation by large corporations has undermined citizen trust and led to an erosion of civic life. By promoting greater transparency and user control, algorithmic pluralism could go a significant way to restoring some of that trust.

- Diversity and Inclusion: A considerable amount of research has pointed to the inherent bias and exclusivity of the algorithms currently used by social media and other platforms.[26][27][28] These biases are reflections of underlying power structures in society and are likely to be exacerbated with a greater reliance on AI-driven algorithms. Algorithmic multiplicity thus has a potentially far greater role to play than simply in promoting informational diversity; by widening the parameters of mediation, bringing in a greater diversity of voices and views, it can help widen our very notions of reality and, perhaps, reduce long-standing structural social and economic inequalities.

## Challenges and Risks:

- Reduced Accountability: While the goal of algorithmic pluralism is to impose greater accountability on tech platforms one of the unintended results may, paradoxically, be the very opposite. Evaluating the prospects of algorithmic choice in 2021, for instance, Robert

---

[26] Lau, Theodora and Akkaraju, Uday. When Algorithms Decide Whose Voices Will Be Heard. Harvard Business Review. https://hbr.org/2019/11/when-algorithms-decide-whose-voice-will-be-heard;

[27] Kulshrestha, Juhi, Motahhare Eslami, Johnnatan Messias, Muhammad Bilal Zafar, Saptarshi Ghosh, Krishna P. Gummadi, and Karrie Karahalios. "Quantifying search bias: Investigating sources of bias for political searches in social media." In Proceedings of the 2017 ACM conference on computer supported cooperative work and social computing, pp. 417-432. 2017.

[28] Schroeder, Jonathan E. "Reinscribing gender: social media, algorithms, bias." Journal of marketing management 37, no. 3-4 (2021): 376-378.

Faris and Joan Donovan[29] argued that requiring platforms to open up their systems could lead them to reduce their existing moderation efforts. An illusion of "choice" would in effect substitute for accountability, reducing legal and reputational incentives to invest in human and technical solutions to limit online racism, radicalism and misinformation.

- User Apathy: Choice may be a good thing, but it's unlikely to be a panacea for the various social ills attached to our current algorithms. Among the key questions proponents of algorithmic pluralism would do well to consider is whether users actually want choice and, if presented with options, how deep they would dig beneath default options. There is a very real possibility that users would choose algorithmic options that simply replicate their existing biases and filter bubbles.

- Lack of Transparency and Explainability: User apathy may not be the only reason for an unwillingness (or inability) to maximize the potential of algorithmic pluralism. A lack of understanding regarding available options could also play a role, as could a lack of clear explainability and transparency on the part of social media platforms. As Faris and Donovan state,[30] "researchers must also consider how effective giving users a choice of filters will be when many may know little about how algorithms curate content in the first place." A main challenge of maximizing the potential of algorithmic pluralism is the unclear or hidden user interface elements that influence user engagement with algorithmic options, which necessitates the efforts to enhance algorithmic literacy. There is a need to move beyond simply offering more algorithmic options to users; greater availability needs to be accompanied by efforts to enhance algorithmic proficiency so that users are more aware of their options and their choices to maximize the potential benefits of algorithms for individuals and society.

---

[29]

[30] Faris, Robert, and Joan Donovan. "The future of platform power: Quarantining misinformation." Journal of Democracy 32, no. 3 (2021): 152-156.

- Market Consolidation: Regulation rarely has a uniform effect across or within industries. The "differential impact"[31] of any given policy or rule can result in unpredictable consequences, sometimes at odds with policymakers' original intent. Requirements for algorithmic pluralism may for instance inadvertently strengthen the most dominant companies or platforms, which would be better placed to meet the resulting technical and financial compliance requirements. This would be a result deeply incongruent with the original animating spirit behind the push for algorithmic pluralism, a desire for greater diversity and competition.

# Conclusion

Much like previous mediating technologies, aIgorithms play a crucial role in shaping our sense of reality. This role resurfaces and reemphasizes the importance of historical policy debates about media dominance and the essential role of pluralism in preserving democratic values. The consolidation of media, and now algorithmic mediation, poses risks to intellectual diversity and democratic discourse, underscoring the need for algorithmic pluralism.

Algorithmic pluralism aims to introduce choice and diversity in AI systems to counter the monopolistic tendencies displayed by tech giants like Meta and OpenAI. It advocates for a market of algorithms where users can choose the biases they interact with, promoting a more nuanced AI interaction. This nascent concept, rooted in addressing the limitations of social media platforms and the "filter bubbles" they create, has begun gaining traction with regulatory frameworks like the EU's Digital Services Act and initiatives like Bluesky.

The potential offered by algorithmic pluralism is vast. It promotes choice, enhances democracy, fosters market competition, encourages innovation, and builds trust through transparency. Additionally, it addresses the critical issue of diversity and inclusion, challenging the inherent biases of existing algorithms. However, it's not devoid of challenges; reduced accountability, user apathy, lack of transparency, and market consolidation are significant hurdles to overcome.

---

[31] Dove, John A. "One size fits all? The differential impact of federal regulation on early-stage entrepreneurial activity across US states." Journal of Regulatory Economics 63, no. 1-2 (2023): 57-73.

In addition there is need to go beyond just providing algorithmic options; enhancing algorithmic literacy, ensuring transparency, and fostering a conducive regulatory environment are also critical for realizing the benefits of algorithmic pluralism. Through a balanced approach that addresses these challenges, algorithmic pluralism can potentially transform the interaction between citizens, AI systems, and the digital world, leading towards a more democratic and inclusive digital society.